\documentclass[runningheads]{llncs}
\usepackage[T1]{fontenc}
\usepackage{microtype}

\usepackage{inconsolata}

\usepackage{multirow}
\usepackage{makecell}

\usepackage{hyperref}

\usepackage{times}
\usepackage{latexsym}
\usepackage{graphicx}
\usepackage{subfigure}
\usepackage{float}
\graphicspath{{images}}
\usepackage{stfloats}
\usepackage{booktabs}
\usepackage{indentfirst}
\usepackage{amsmath}
\usepackage{amssymb}
\usepackage{mathtools}

\usepackage{enumitem}

\usepackage{colortbl}
\usepackage{arydshln}

\usepackage[linesnumbered,ruled,vlined]{algorithm2e}

\SetCommentSty{mycommfont}

\SetKwInput{KwInput}{Input}                
\SetKwInput{KwOutput}{Output}              

\usepackage{listings}
\usepackage{xcolor}

\definecolor{codegreen}{rgb}{0,0.6,0}
\definecolor{codegray}{rgb}{0.5,0.5,0.5}
\definecolor{codepurple}{rgb}{0.58,0,0.82}
\definecolor{backcolour}{rgb}{0.95,0.95,0.92}

\lstdefinestyle{mystyle}{
    backgroundcolor=\color{backcolour},   
    commentstyle=\color{codegreen},
    keywordstyle=\color{magenta},
    numberstyle=\tiny\color{codegray},
    stringstyle=\color{codepurple},
    basicstyle=\ttfamily\small,
    breakatwhitespace=false,         
    breaklines=true,                 
    captionpos=b,                    
    keepspaces=true,                 
    numbers=left,                    
    numbersep=5pt,                  
    showspaces=false,                
    showstringspaces=false,
    showtabs=false,                  
    tabsize=2
}

\usepackage{CJKutf8}

\begin{document}

\begin{CJK*}{UTF8}{gbsn}

\title{Diffu-LoRA: A Novel Low-Rank Adaptation for Personalized Diffusion Models}
%
%
\author{Tianjing Li\inst{1}\thanks{Corresponding: \email{liz\_li@dataselect21.com}. } 
\and
Wei Zhu\inst{1}
}
\authorrunning{Li et al.}
%
\institute{Zhangjiang Lab of Artificial Intelligence, Shanghai, China }

\maketitle              
\begin{abstract}
Personalizing text-to-image diffusion models from a few reference images requires preserving subject identity while following prompts that describe new contexts. Full-model fine-tuning is parameter-intensive, whereas low-rank adaptation (LoRA) reduces the number of trainable parameters but leaves open how adaptation capacity should be distributed across layers. We introduce Diffu-LoRA, a parameter-efficient method that learns this allocation through gated low-rank adaptation. Diffu-LoRA inserts trainable low-rank components into the linear layers of Transformer blocks and assigns a learnable gate to each component. Bilevel optimization updates the adaptation weights and gate parameters on separate data splits, while progressive pruning removes components with the lowest gate values to meet a prescribed rank budget. This procedure allocates adaptation capacity nonuniformly across layers while keeping the pretrained backbone frozen. Experiments with Stable Diffusion on subjects from DreamBooth and additional collected datasets show improved overall subject fidelity and prompt alignment relative to the evaluated fine-tuning baselines. Ablation studies examine the contributions of bilevel optimization, progressive pruning, and adapter placement. These results support learned rank allocation as a practical approach to parameter-efficient diffusion model personalization.

\end{abstract}

\section{Introduction}

Text-to-image (T2I) diffusion models can generate diverse images from natural-language descriptions, but their ability to depict a particular subject is limited when that subject is absent from the training data or difficult to specify in words. Subject-driven generation addresses this limitation by conditioning or adapting a pretrained model on a small collection of reference images~\cite{shuai2024survey}. The aim is to preserve the subject's distinctive appearance while allowing changes in pose, background, and artistic style. Methods such as DreamBooth~\cite{dreamBooth} and BLIP-Diffusion~\cite{li2024blip} have made this setting practical, building on the generative capabilities of pretrained diffusion models, including latent diffusion models~\cite{stable_diffusion}. Nevertheless, reproducing a subject in unfamiliar contexts remains difficult: a model must learn which visual features define the subject without treating incidental details in the reference images as part of its identity.

Fine-tuning offers a direct way to incorporate a new subject, but updating a large diffusion backbone for each subject is costly in both optimization and storage. With only a few reference images, adaptation may also overfit to the observed backgrounds or poses, compromising the response to new prompts. Parameter-efficient fine-tuning (PEFT) reduces the adaptation cost by optimizing a small set of parameters while freezing the pretrained backbone~\cite{Ding2022DeltaTA}. In particular, low-rank adaptation (LoRA) represents weight updates through trainable low-rank factors~\cite{hu2021lora}. Its effectiveness, however, depends not only on the total adaptation capacity but also on where that capacity is placed. Assigning the same rank to selected layers is convenient, yet it does not account for their potentially different roles in learning subject-specific features and maintaining prompt responsiveness. This raises a practical question: how should a limited rank budget be distributed across the diffusion model?

We address this question with Diffu-LoRA, a parameter-efficient method that learns the allocation of low-rank adaptation capacity during personalization. Diffu-LoRA equips the linear layers of Transformer blocks with gated low-rank updates. For an adapted layer $\ell$, the update takes the form
\begin{equation}
    \Delta W_{\ell}
    = U_{\ell}\operatorname{diag}(\mathbf{g}_{\ell})V_{\ell},
    \label{eq:diffu_lora_update}
\end{equation}
where
$U_{\ell}\in\mathbb{R}^{d_{\ell}^{\mathrm{in}}\times q_{\ell}}$
and
$V_{\ell}\in\mathbb{R}^{q_{\ell}\times d_{\ell}^{\mathrm{out}}}$
are trainable factors, $q_{\ell}$ is the initial adapter rank, and
$\mathbf{g}_{\ell}\in\mathbb{R}^{q_{\ell}}$ contains component-wise gates.
The pretrained weights remain frozen. We formulate rank allocation as a bilevel optimization problem: the low-rank factors are fitted on one data split, whereas the gate parameters are updated on a separate split. Gate values provide a learned criterion for progressively removing low-rank components until the prescribed rank budget is reached. Adaptation continues between pruning stages, allowing the remaining components to adjust to the reduced capacity. The resulting allocation can vary across layers rather than imposing a uniform rank throughout. We evaluate Diffu-LoRA with Stable Diffusion on subjects from DreamBooth and additional collected datasets, examining both subject fidelity and alignment with the requested prompts.

Our contributions are summarized as follows:
\begin{itemize}
    \item We introduce Diffu-LoRA, which combines component-wise gating,
    bilevel optimization, and progressive pruning to learn a nonuniform
    allocation of low-rank adaptation capacity for diffusion model
    personalization under a prescribed rank budget.
    \item We evaluate Diffu-LoRA against the considered fine-tuning
    baselines and observe improved overall subject fidelity and prompt
    alignment. Ablation studies examine the effects of bilevel
    optimization, progressive pruning, and adapter placement.
\end{itemize}

\section{Related Work}
\label{sec:related_work}

\subsection{Parameter-Efficient Fine-Tuning}
\label{sec:related_peft}

Parameter-efficient fine-tuning (PEFT) adapts a pretrained model by
optimizing only a small set of parameters while leaving most, or all, of
the backbone frozen~\cite{Ding2022DeltaTA}. Existing approaches can be
broadly grouped according to how the trainable parameters are introduced.
Adapter-based methods insert compact neural modules into selected layers
of the backbone~\cite{houlsby2019parameter,Rckl2020AdapterDropOT}, whereas
prefix tuning and prompt tuning learn continuous conditioning vectors
without modifying the original model weights
directly~\cite{li2021prefix,lester2021power,Liu2022PTuningPT}. Other
methods restrict optimization to a predefined subset of parameters or
learn masks that identify which parameters should be updated
~\cite{BenZaken2021BitFitSP,guo-etal-2021-parameter,zhao-etal-2020-masking}.
Multiplicative schemes such as (IA)$^{3}$ instead modulate intermediate
activations through a small number of learned scaling vectors
~\cite{Liu2022FewShotPF}. Reparameterization-based methods constitute
another important family: rather than updating a dense weight matrix,
they express its task-specific change through a compact parameterization.
Their design is supported by evidence that adaptation often occurs in a
low-dimensional subspace~\cite{aghajanyan-etal-2021-intrinsic}.

The storage advantage of PEFT is particularly relevant to personalized
text-to-image generation, where a separate adaptation is commonly
maintained for each subject. Full fine-tuning methods such as
DreamBooth~\cite{dreamBooth} can capture subject-specific appearance but
require substantially more trainable parameters. Diffu-LoRA instead
retains a shared frozen diffusion backbone and learns compact,
subject-specific low-rank updates. Unlike conventional PEFT schemes with
a fixed capacity at every adapted layer, it determines how a prescribed
global rank budget should be distributed across the denoiser.

\subsection{Low-Rank Adaptation and Its Variants}
\label{sec:related_lora}

Low-rank adaptation (LoRA) represents the update to a frozen weight matrix
as the product of two trainable low-rank factors~\cite{hu2021lora}. This
parameterization substantially reduces both optimization and per-task
storage costs and has consequently motivated several complementary
extensions. QLoRA combines low-rank adaptation with quantized backbone
weights to reduce the memory required during fine-tuning
~\cite{2023arXiv230514314D}. LoRA-FA lowers activation-memory consumption
by freezing one of the two low-rank factors and optimizing only the other
~\cite{Zhang2023LoRAFAML}. VeRA further reduces the number of learned
parameters by sharing frozen random projection matrices across layers and
training lightweight scaling vectors~\cite{Kopiczko2023VeRAVR}. These
methods primarily improve the memory or parameter efficiency of a
predefined LoRA configuration. Diffu-LoRA addresses a different but
complementary question: given a fixed total capacity, which low-rank
components should be retained in each adapted layer? It learns
component-wise gates on a held-out split and progressively removes
low-scoring components, producing layer-dependent ranks without updating
the pretrained diffusion backbone.

\section{Preliminaries}
\label{sec:preliminaries}

\subsection{Bilevel Optimization}
\label{sec:bilevel_optimization}

Bilevel optimization separates the learning process into an inner problem and an outer problem~\cite{Liu2019DARTSDA}. Let $\boldsymbol{\Omega}$ denote the model parameters optimized at the inner level and let $\boldsymbol{\theta}$ denote the variables controlled by the outer level. Given two disjoint data splits, $\mathcal{D}_{\mathrm{tr}}$ and $\mathcal{D}_{\mathrm{val}}$, the corresponding objective is
\begin{equation}
\begin{aligned}
    \min_{\boldsymbol{\theta}}\quad
    & \mathcal{L}_{\mathrm{val}}
      \bigl(\boldsymbol{\Omega}^{*}(\boldsymbol{\theta}),
            \boldsymbol{\theta}\bigr), \\
    \text{s.t.}\quad
    & \boldsymbol{\Omega}^{*}(\boldsymbol{\theta})
      \in
      \arg\min_{\boldsymbol{\Omega}}
      \mathcal{L}_{\mathrm{tr}}
      \bigl(\boldsymbol{\Omega},\boldsymbol{\theta}\bigr),
\end{aligned}
\label{eq:bilevel_problem}
\end{equation}
where $\mathcal{L}_{\mathrm{tr}}$ and $\mathcal{L}_{\mathrm{val}}$ are evaluated on $\mathcal{D}_{\mathrm{tr}}$ and $\mathcal{D}_{\mathrm{val}}$, respectively. In practice, solving the inner problem to convergence after every outer update is prohibitively expensive. We therefore use alternating gradient updates: $\boldsymbol{\Omega}$ is updated using samples from $\mathcal{D}_{\mathrm{tr}}$, whereas $\boldsymbol{\theta}$ is updated using samples from $\mathcal{D}_{\mathrm{val}}$. This separation discourages the outer variables from merely fitting the same examples used to optimize the model parameters.

\subsection{CLIP}
\label{sec:clip}

Contrastive Language--Image Pre-training (CLIP) learns a shared representation space for images and natural-language descriptions from paired image--text data~\cite{radford2021learning}. Let $f_{\mathrm{I}}$ and $f_{\mathrm{T}}$ denote the CLIP image and text encoders, respectively. For an image $\mathbf{x}$ and a text prompt $p$, their normalized representations are
\begin{equation}
\begin{aligned}
    \bar{\mathbf{v}}(\mathbf{x})
    &=
    \frac{f_{\mathrm{I}}(\mathbf{x})}
         {\lVert f_{\mathrm{I}}(\mathbf{x}) \rVert_{2}}, \\
    \bar{\mathbf{u}}(p)
    &=
    \frac{f_{\mathrm{T}}(p)}
         {\lVert f_{\mathrm{T}}(p) \rVert_{2}} .
\end{aligned}
\label{eq:clip_embeddings}
\end{equation}
Their semantic agreement can then be measured by
$s_{\mathrm{CLIP}}(\mathbf{x},p)
=\bar{\mathbf{v}}(\mathbf{x})^{\mathsf{T}}\bar{\mathbf{u}}(p)$.
Because both representations have unit norm, this inner product is equivalent to cosine similarity. In this work, CLIP embeddings are used to quantify both visual similarity and prompt--image alignment.

\subsection{Text-to-Image Diffusion Models}
\label{sec:text_to_image_diffusion}

A text-to-image latent diffusion model learns to reverse a gradual noising process in a compact latent space~\cite{stable_diffusion}. Given an image $\mathbf{x}$, a variational autoencoder maps it to a clean latent representation $\mathbf{z}_{0}=\mathcal{E}(\mathbf{x})$. At diffusion step $t$, a noisy latent is constructed as
$\mathbf{z}_{t}=\alpha_{t}\mathbf{z}_{0}
+\sigma_{t}\boldsymbol{\epsilon}$, where
$\boldsymbol{\epsilon}\sim\mathcal{N}(\mathbf{0},\mathbf{I})$
and $\alpha_{t}$ and $\sigma_{t}$ are determined by the noise schedule. A text encoder $\tau$ maps the prompt $p$ to a conditioning representation $\mathbf{c}=\tau(p)$. The denoising network $\boldsymbol{\epsilon}_{\phi}$ is trained with the weighted noise-prediction objective
\begin{equation}
    \mathcal{L}_{\mathrm{diff}}(\phi)
    =
    \mathbb{E}_{\mathbf{x},p,t,\boldsymbol{\epsilon}}
    \left[
        w(t)
        \left\lVert
            \boldsymbol{\epsilon}
            -
            \boldsymbol{\epsilon}_{\phi}
            (\mathbf{z}_{t},t,\mathbf{c})
        \right\rVert_{2}^{2}
    \right],
\label{eq:diffusion_objective}
\end{equation}
where $w(t)$ is an optional timestep-dependent weighting function. During generation, the reverse process starts from Gaussian noise and repeatedly applies the conditional denoiser; the resulting latent is subsequently decoded by $\mathcal{E}$'s paired decoder. Personalization adapts a subset of the denoiser parameters while retaining the generative knowledge of the pretrained model.

\subsection{Low-Rank Adaptation}
\label{sec:lora_preliminaries}

Low-rank adaptation (LoRA) is motivated by the observation that the parameter changes required for downstream adaptation often lie in a low-dimensional subspace~\cite{aghajanyan2020intrinsic,hu2021lora}. Instead of updating a pretrained weight matrix directly, LoRA represents its task-specific modification by two trainable low-rank factors. The original model parameters remain frozen, substantially reducing the number of parameters that must be optimized and stored for each personalized subject.

Consider a linear layer $\ell$ with frozen weight
$\mathbf{W}_{\ell}^{0}
\in\mathbb{R}^{d_{\ell}^{\mathrm{in}}
\times d_{\ell}^{\mathrm{out}}}$
and bias $\mathbf{b}_{\ell}$. Under the row-vector convention used throughout this paper, LoRA modifies its forward map as
\begin{equation}
\begin{aligned}
    \mathbf{y}_{\ell}
    &=
    \mathbf{x}_{\ell}
    \bigl(\mathbf{W}_{\ell}^{0}
          +\Delta\mathbf{W}_{\ell}\bigr)
    +\mathbf{b}_{\ell}, \\
    \Delta\mathbf{W}_{\ell}
    &=
    \mathbf{U}_{\ell}\mathbf{V}_{\ell},
\end{aligned}
\label{eq:lora_update}
\end{equation}
where
$\mathbf{U}_{\ell}
\in\mathbb{R}^{d_{\ell}^{\mathrm{in}}\times r_{\ell}}$
and
$\mathbf{V}_{\ell}
\in\mathbb{R}^{r_{\ell}\times d_{\ell}^{\mathrm{out}}}$
are trainable, with
$r_{\ell}\ll
\min(d_{\ell}^{\mathrm{in}},d_{\ell}^{\mathrm{out}})$.
Consequently,
$\operatorname{rank}(\Delta\mathbf{W}_{\ell})\leq r_{\ell}$,
and the adapter contains
$r_{\ell}(d_{\ell}^{\mathrm{in}}+d_{\ell}^{\mathrm{out}})$
parameters rather than
$d_{\ell}^{\mathrm{in}}d_{\ell}^{\mathrm{out}}$.
Standard LoRA typically assigns ranks in advance, whereas Diffu-LoRA learns how the available low-rank components should be distributed across the adapted layers.

\section{Methodology}
\label{sec:methodology}

\subsection{Problem Formulation}
\label{sec:problem_formulation}

\noindent\textbf{Diffusion model personalization.}
Given a small set of reference images depicting a subject, our goal is to adapt a pretrained text-to-image diffusion model so that it can reproduce the subject under previously unseen prompts. Let $\phi$ denote the frozen parameters of the pretrained denoiser and let $\mathcal{D}_{\mathrm{ins}}$ be the instance dataset. Personalization should preserve the visual characteristics shared by the reference images while avoiding incidental properties, such as a particular background, viewpoint, or pose. Following DreamBooth~\cite{dreamBooth}, we formulate this task as conditional diffusion training rather than direct image reconstruction.

\noindent\textbf{Prompts for few-shot personalization.}
Each reference image is paired with an instance prompt of the form ``a \texttt{[V]} \texttt{[C]},'' where \texttt{[V]} is a rare identifier assigned to the subject and \texttt{[C]} is its coarse semantic class, such as \emph{dog}, \emph{backpack}, or \emph{watch}. The identifier distinguishes the target subject from other members of the same class, whereas the class noun connects it to the semantic knowledge already encoded by the pretrained model. At inference time, \texttt{[V]} can be combined with novel descriptions of scenes, attributes, actions, and artistic styles.

\noindent\textbf{Prior-preserving objective.}
Fine-tuning on only a few instance images may bind the class noun too closely to the target subject and reduce the diversity of generated poses or viewpoints. We therefore construct a class-prior set $\mathcal{D}_{\mathrm{pr}}$ by sampling the frozen model with the class prompt ``a \texttt{[C]}.'' For an image--prompt pair $(\mathbf{x},p)$, define the adapted diffusion loss as
\begin{equation}
\begin{split}
    \ell(\mathbf{x},p;\boldsymbol{\Omega},\boldsymbol{a})
    =
    \mathbb{E}_{t,\boldsymbol{\epsilon}}
    \Bigl[
        w(t)
        \bigl\|
            \boldsymbol{\epsilon}
            -
            \boldsymbol{\epsilon}_{\phi,\boldsymbol{\Omega},
            \boldsymbol{a}}
            (\mathbf{z}_{t},t,\tau(p))
        \bigr\|_{2}^{2}
    \Bigr],
    \qquad
    \mathbf{z}_{t}
    =
    \alpha_{t}\mathcal{E}(\mathbf{x})
    +
    \sigma_{t}\boldsymbol{\epsilon},
\end{split}
\label{eq:personalization_sample_loss}
\end{equation}
where $\boldsymbol{\Omega}$ contains the low-rank factors and
$\boldsymbol{a}$ contains their gate parameters. For any data split
$\mathcal{S}=(\mathcal{S}_{\mathrm{ins}},\mathcal{S}_{\mathrm{pr}})$,
the complete personalization objective is
\begin{equation}
\begin{split}
    \mathcal{J}
    (\mathcal{S};\boldsymbol{\Omega},\boldsymbol{a})
    =
    \frac{1}{|\mathcal{S}_{\mathrm{ins}}|}
    \sum_{(\mathbf{x},p)\in\mathcal{S}_{\mathrm{ins}}}
    \ell(\mathbf{x},p;\boldsymbol{\Omega},\boldsymbol{a})
    +
    \frac{\lambda_{\mathrm{pr}}}
         {|\mathcal{S}_{\mathrm{pr}}|}
    \sum_{(\mathbf{x},p)\in\mathcal{S}_{\mathrm{pr}}}
    \ell(\mathbf{x},p;\boldsymbol{\Omega},\boldsymbol{a}),
\end{split}
\label{eq:personalization_objective}
\end{equation}
where $\lambda_{\mathrm{pr}}\geq 0$ controls prior preservation.

\noindent\textbf{Parameter-efficient adaptation.}
Maintaining a separate copy of the entire diffusion model for every subject is impractical in a multi-user system. Diffu-LoRA instead freezes the shared backbone and stores only a compact set of subject-specific adapter parameters. A single pretrained model can therefore serve multiple subjects by loading the corresponding adapters on demand. The adapters may either remain separate from the backbone, as in multi-tenant inference systems~\cite{Chen2023PunicaML}, or be merged into the associated linear weights when a dedicated model instance is required.

Formally, let $P_{\phi}$ be the number of frozen backbone parameters and let $r_{\ell}$ denote the final adapter rank assigned to an adapted layer $\ell$. The number of subject-specific factor parameters is
\begin{equation}
    P_{\mathrm{ad}}
    =
    \sum_{\ell\in\mathcal{A}}
    r_{\ell}
    \left(
        d_{\ell}^{\mathrm{in}}
        +
        d_{\ell}^{\mathrm{out}}
    \right),
    \qquad
    P_{\mathrm{ad}}\ll P_{\phi},
\label{eq:adapter_parameter_count}
\end{equation}
where $\mathcal{A}$ is the set of adapted linear layers. The objective is thus to minimize Eq.~\eqref{eq:personalization_objective} while selecting the ranks $\{r_{\ell}\}_{\ell\in\mathcal{A}}$ under a prescribed global budget. This formulation separates the reusable generative knowledge in $\phi$ from the small amount of information required to represent a particular subject.

\subsection{Gated Low-Rank Adaptation}
\label{sec:gated_lora}

Diffu-LoRA inserts an overparameterized low-rank adapter into each linear layer in $\mathcal{A}$, which includes the linear transformations of the selected Transformer blocks. Layer $\ell$ is initialized with $q_{\ell}$ candidate rank-one components, giving an initial component count
$B_{0}=\sum_{\ell\in\mathcal{A}}q_{\ell}$.
The desired model must satisfy
\begin{equation}
    \sum_{\ell\in\mathcal{A}} r_{\ell}
    =
    B_{\mathrm{tar}},
    \qquad
    B_{\mathrm{tar}}<B_{0},
\label{eq:global_rank_budget}
\end{equation}
where $B_{\mathrm{tar}}$ is the target rank budget. In contrast to assigning the same rank to every layer, Diffu-LoRA learns the individual values of $r_{\ell}$ from the personalization data.

For layer $\ell$, let
$\mathbf{U}_{\ell}\in
\mathbb{R}^{d_{\ell}^{\mathrm{in}}\times q_{\ell}}$
and
$\mathbf{V}_{\ell}\in
\mathbb{R}^{q_{\ell}\times d_{\ell}^{\mathrm{out}}}$
be its low-rank factors. Each candidate component has a learnable logit
$a_{\ell i}\in\mathbb{R}$ and a binary activity mask
$m_{\ell i}\in\{0,1\}$. Its effective gate is defined by
$g_{\ell i}=2m_{\ell i}\sigma(a_{\ell i})$, where $\sigma(\cdot)$ is the logistic sigmoid. Under the row-vector convention of Sect.~\ref{sec:lora_preliminaries}, the adapted layer is
\begin{equation}
\begin{split}
    \mathbf{y}_{\ell}
    &=
    \mathbf{x}_{\ell}
    \left(
        \mathbf{W}_{\ell}^{0}
        +
        \mathbf{U}_{\ell}
        \operatorname{diag}(\mathbf{g}_{\ell})
        \mathbf{V}_{\ell}
    \right)
    +\mathbf{b}_{\ell},                                      \\
    \mathbf{g}_{\ell}
    &=
    2\mathbf{m}_{\ell}
    \odot
    \sigma(\mathbf{a}_{\ell}),
\end{split}
\label{eq:diffu_lora_layer}
\end{equation}
where $\odot$ denotes element-wise multiplication. Initially,
$\mathbf{m}_{\ell}=\mathbf{1}$ and
$\mathbf{a}_{\ell}=\mathbf{0}$, so every gate equals one. We initialize one low-rank factor to zero, ensuring that the adapter produces no perturbation before optimization begins.

The factor parameters
$\boldsymbol{\Omega}
=\{\mathbf{U}_{\ell},\mathbf{V}_{\ell}\}_{\ell\in\mathcal{A}}$
and gate logits
$\boldsymbol{a}
=\{\mathbf{a}_{\ell}\}_{\ell\in\mathcal{A}}$
play different roles: the former encode the subject, whereas the latter determine where adaptation capacity is retained. We consequently divide the available personalization data into disjoint splits
$\mathcal{D}_{\mathrm{tr}}$ and $\mathcal{D}_{\mathrm{val}}$
and optimize
\begin{equation}
\begin{aligned}
    \min_{\boldsymbol{a}}\quad
    &\mathcal{J}
    \bigl(
        \mathcal{D}_{\mathrm{val}};
        \boldsymbol{\Omega}^{*}(\boldsymbol{a}),
        \boldsymbol{a}
    \bigr),                                                   \\
    \text{s.t.}\quad
    &\boldsymbol{\Omega}^{*}(\boldsymbol{a})
    \in
    \arg\min_{\boldsymbol{\Omega}}
    \mathcal{J}
    \bigl(
        \mathcal{D}_{\mathrm{tr}};
        \boldsymbol{\Omega},
        \boldsymbol{a}
    \bigr).
\end{aligned}
\label{eq:diffu_lora_bilevel}
\end{equation}
As in Sect.~\ref{sec:bilevel_optimization}, we approximate this problem using alternating first-order updates: low-rank factors are updated on $\mathcal{D}_{\mathrm{tr}}$, while gate logits are updated on $\mathcal{D}_{\mathrm{val}}$. The held-out objective makes the gates less likely to favor components that merely memorize the reference images.

\subsection{Progressive Rank Pruning}
\label{sec:lora_pruning}

After the initial bilevel warm-up, Diffu-LoRA progressively reduces the number of active components from $B_{0}$ to $B_{\mathrm{tar}}$. At each pruning stage, active components are ranked by their current gate values $g_{\ell i}$, and the $n_{0}$ lowest-scoring components are removed by setting their masks to zero. Their masks and gate logits remain frozen thereafter. The surviving factors are then optimized for an additional $K_{2}$ steps before the next pruning stage, allowing them to compensate for the lost components. Once the target budget is reached, inactive columns of $\mathbf{U}_{\ell}$ and the corresponding rows of $\mathbf{V}_{\ell}$ are physically discarded. The remaining gates can be absorbed into either factor, leaving a standard low-rank update with layer-dependent rank $r_{\ell}$. Algorithm~\ref{algo:diffu_lora} summarizes the complete procedure.

\begin{algorithm}[t]
\small
\DontPrintSemicolon

\KwInput{Frozen denoiser parameters $\phi$;
adapted layers $\mathcal{A}$;
data splits $\mathcal{D}_{\mathrm{tr}}$ and
$\mathcal{D}_{\mathrm{val}}$}

\KwInput{Initial ranks $\{q_{\ell}\}$;
target budget $B_{\mathrm{tar}}$;
pruning size $n_{0}$;
intervals $K_{1}$ and $K_{2}$}

\KwOutput{Compacted Diffu-LoRA factors
$\{\mathbf{U}_{\ell},\mathbf{V}_{\ell}\}_{\ell\in\mathcal{A}}$}

Initialize $\boldsymbol{\Omega}$, set
$\mathbf{a}_{\ell}\leftarrow\mathbf{0}$ and
$\mathbf{m}_{\ell}\leftarrow\mathbf{1}$ for all
$\ell\in\mathcal{A}$\;

\For{$k\leftarrow 1$ \KwTo $K_{1}$}{
    Update $\boldsymbol{\Omega}$ using
    $\mathcal{D}_{\mathrm{tr}}$\;
    Update $\boldsymbol{a}$ using
    $\mathcal{D}_{\mathrm{val}}$\;
}

\While{$\sum_{\ell}\|\mathbf{m}_{\ell}\|_{0}
       > B_{\mathrm{tar}}$}{
    $h\leftarrow
    \min\!\left\{
        n_{0},
        \sum_{\ell}\|\mathbf{m}_{\ell}\|_{0}
        -B_{\mathrm{tar}}
    \right\}$\;

    Select the $h$ active components with the
    smallest $g_{\ell i}$\;

    Set their masks to zero and freeze their logits\;

    \For{$k\leftarrow 1$ \KwTo $K_{2}$}{
        Update $\boldsymbol{\Omega}$ using
        $\mathcal{D}_{\mathrm{tr}}$\;
        Update active entries of $\boldsymbol{a}$ using
        $\mathcal{D}_{\mathrm{val}}$\;
    }
}

Remove inactive factor entries and absorb the
surviving gates into the low-rank factors\;

\caption{Training and rank allocation in Diffu-LoRA.}
\label{algo:diffu_lora}
\end{algorithm}

\section{Experiments}
\label{sec:experiments}

We evaluate Diffu-LoRA on subject-driven image generation using the publicly available Stable Diffusion backbone~\cite{stable_diffusion}. The experiments are designed to examine both the overall personalization performance and the contribution of each component introduced in Sect.~\ref{sec:methodology}. We compare Diffu-LoRA with full fine-tuning and conventional LoRA configurations under the same DreamBooth training objective, and then study the effects of bilevel optimization, progressive pruning, and adapter placement.

\subsection{Datasets and Evaluation Protocol}
\label{sec:datasets}

Our main benchmark consists of 12 subjects from the DreamBooth dataset~\cite{dreamBooth}, covering animals and objects such as dogs, cats, backpacks, toys, and household items. We also include six subjects from Custom Diffusion~\cite{kumari2023multi} to increase the diversity of object categories and visual appearances. The rare subject identifiers and class nouns used during personalization follow the settings of the corresponding original datasets.

To evaluate more challenging visual concepts, we additionally collect reference images for human faces, anime characters, and distinctive objects. Each face subject contains four to seven images captured from different viewpoints, allowing us to examine whether a method preserves identity-related spatial features. The complete evaluation set is organized into four groups: living subjects, inanimate objects, anime characters, and human faces. Representative reference images are shown in Fig.~\ref{fig:dreambooth_datasets}. The sources of the six Internet-collected subjects reported in Table~\ref{tab:results_main_2} are listed below.
\begin{itemize}[leftmargin=*,itemsep=2pt,topsep=3pt]
    \item \textbf{Rolls-Royce:}
    \begin{itemize}[leftmargin=*,itemsep=1pt,topsep=1pt]
        \item \url{https://www.peakpx.com/en/hd-wallpaper-desktop-pxxec}
        \item \url{https://4kwallpapers.com/cars/rolls-royce-ghost-2020/white-background-5k-8k-2554.html}
        \item \url{https://www.cardekho.com/Rolls-Royce/Rolls-Royce_Ghost/pictures#leadForm}
        \item \url{https://www.rolls-roycemotorcars.com/en_US/showroom/ghost-digital-brochure.html}
    \end{itemize}

    \item \textbf{Hugging Face:}
    \url{https://huggingface.co/brand}

    \item \textbf{Nami:}
    \begin{itemize}[leftmargin=*,itemsep=1pt,topsep=1pt]
        \item \url{http://m.gettywallpapers.com/nami-pfps-2/}
        \item \url{https://tensor.art/models/616615209278282245}
        \item \url{https://www.facebook.com/NamiHotandCute/?locale=bs_BA}
        \item \url{https://k.sina.cn/article_1655152542_p62a79f9e02700nhhe.html}
    \end{itemize}

    \item \textbf{Kiriko:}
    \begin{itemize}[leftmargin=*,itemsep=1pt,topsep=1pt]
        \item \url{https://in.pinterest.com/pin/306948530865002366/}
        \item \url{https://encrypted-tbn2.gstatic.com/images?q=tbn:ANd9GcSDSk98Uw3O2XW_RFC1jD_Kmw70JWU459euVYtU9nn1CpzPDcwS}
        \item \url{https://comisc.theothertentacle.com/overwatch+kiriko+fanart}
        \item \url{https://www.1999.co.jp/eng/11030018}
    \end{itemize}

    \item \textbf{Shoko Komi:}
    \begin{itemize}[leftmargin=*,itemsep=1pt,topsep=1pt]
        \item \url{https://wallpaperforu.com/tag/komi-shouko-wallpaper/page/2/}
        \item \url{https://www.tiktok.com/@anime_geek00/video/7304798157894995243}
        \item \url{https://wall.alphacoders.com/big.php?i=1305702}
        \item \url{http://m.gettywallpapers.com/komi-can-t-communicate-wallpapers/}
    \end{itemize}

    \item \textbf{Kakashi Hatake:}
    \begin{itemize}[leftmargin=*,itemsep=1pt,topsep=1pt]
        \item \url{https://www.ranker.com/list/best-kakashi-hatake-quotes/ranker-anime?page=2}
        \item \url{https://www.wallpaperflare.com/search?wallpaper=Hatake+Kakashi}
        \item \url{https://www.peakpx.com/en/hd-wallpaper-desktop-kiptm}
        \item \url{https://in.pinterest.com/pin/584060645404620659/}
    \end{itemize}
\end{itemize}

\subsection{Implementation Details}
\label{sec:implementation_details}

All methods use Stable Diffusion v1.5 as the pretrained backbone. Unless otherwise stated, the backbone parameters are frozen and only the parameters specified by each adaptation method are optimized. The principal implementation settings are summarized below.
\begin{itemize}[leftmargin=*,itemsep=3pt,topsep=3pt]
    \item We use the publicly available Stable Diffusion v1.5 checkpoint.\footnote{\url{https://huggingface.co/runwayml/stable-diffusion-v1-5}}

    \item Each selected Transformer block contains six adapted linear transformations, including the attention projections and feed-forward transformations. Every adapted layer is initialized with
    $q_{\ell}=32$ candidate components. Thus, the initial component budget of each block is
    $B_{0}=6\times32=192$, and the target budget is set to
    $B_{\mathrm{tar}}=96$. Rank allocation is performed independently but concurrently for all selected blocks.

    \item Personalization is run for 200 optimization steps with a batch size of one. We use AdamW~\cite{Loshchilov2019DecoupledWD}, an initial learning rate of
    $5\times10^{-4}$, and a linear learning-rate decay schedule. The bilevel warm-up lasts
    $K_{1}=20$ steps. We subsequently prune
    $n_{0}=8$ components from each allocation unit at every stage and optimize the surviving components for
    $K_{2}=15$ steps. Twelve pruning stages reduce the per-block budget from 192 to 96 components.

    \item Training and evaluation are conducted on a single NVIDIA GeForce RTX 3090 GPU with 24\,GB of memory.

    \item Hyperparameters not specified above follow the corresponding Hugging Face implementation~\cite{hug_transformers} and are kept fixed across the compared methods.
\end{itemize}

\subsection{Evaluation Metrics}
\label{sec:evaluation_metrics}

We report DINO, CLIP-I, and CLIP-T, with larger values indicating better performance. DINO and CLIP-I measure subject fidelity, whereas CLIP-T measures agreement between a generated image and its conditioning prompt. Let
$\mathcal{R}=\{\mathbf{x}_{i}^{\mathrm{ref}}\}_{i=1}^{n}$
be the reference set and let
$\mathcal{G}=\{(\hat{\mathbf{x}}_{j},p_{j})\}_{j=1}^{m}$
contain the generated images and their corresponding prompts. We use the normalized CLIP representations
$\bar{\mathbf{v}}(\cdot)$ and $\bar{\mathbf{u}}(\cdot)$ defined in Eq.~\eqref{eq:clip_embeddings}. For the DINO encoder
$f_{\mathrm{D}}$, we similarly define
\begin{equation}
    \bar{\mathbf{d}}(\mathbf{x})
    =
    \frac{f_{\mathrm{D}}(\mathbf{x})}
         {\left\lVert f_{\mathrm{D}}(\mathbf{x})\right\rVert_{2}} .
    \label{eq:dino_embedding}
\end{equation}
The three evaluation scores are then written as
\begin{equation}
\begin{aligned}
    S_{\mathrm{CLIP\text{-}I}}
    &=
    \frac{1}{nm}
    \sum_{i=1}^{n}\sum_{j=1}^{m}
    \bar{\mathbf{v}}(\mathbf{x}_{i}^{\mathrm{ref}})^{\mathsf T}
    \bar{\mathbf{v}}(\hat{\mathbf{x}}_{j}), \\[2pt]
    S_{\mathrm{CLIP\text{-}T}}
    &=
    \frac{1}{m}
    \sum_{j=1}^{m}
    \bar{\mathbf{v}}(\hat{\mathbf{x}}_{j})^{\mathsf T}
    \bar{\mathbf{u}}(p_{j}), \\[2pt]
    S_{\mathrm{DINO}}
    &=
    \frac{1}{nm}
    \sum_{i=1}^{n}\sum_{j=1}^{m}
    \bar{\mathbf{d}}(\mathbf{x}_{i}^{\mathrm{ref}})^{\mathsf T}
    \bar{\mathbf{d}}(\hat{\mathbf{x}}_{j}).
\end{aligned}
\label{eq:evaluation_metrics}
\end{equation}

The three metrics capture complementary aspects of personalization. DINO is particularly sensitive to differences among subjects belonging to the same semantic class, while CLIP-I measures broader visual similarity. Neither score should be interpreted without CLIP-T: a model that reproduces the reference images may obtain high subject-fidelity scores while responding poorly to novel prompts. Conversely, the unadapted text-to-image model may follow a prompt well but fail to depict the designated subject. We therefore assess the methods according to the joint behavior of all three metrics.

\subsection{Compared Methods}
\label{sec:compared_methods}

The comparison focuses on adaptation strategies trained with the same DreamBooth instance and prior-preservation objectives. This design isolates the effect of parameterization and adapter placement from that of the personalization loss. We evaluate the following methods:
\begin{itemize}[leftmargin=*,itemsep=3pt,topsep=3pt]
    \item \textbf{DreamBooth + FT} updates all trainable parameters in the Transformer blocks of the Stable Diffusion denoiser.

    \item \textbf{DreamBooth + LoRA-Attn} inserts conventional LoRA adapters~\cite{hu2021lora} only into the linear projections of the multi-head attention modules.

    \item \textbf{DreamBooth + LoRA-All} applies conventional LoRA adapters to all six linear transformations in each selected Transformer block.

    \item \textbf{DreamBooth + Diffu-LoRA} applies the proposed gated adapters to all selected linear transformations and learns their layer-dependent ranks using bilevel optimization and progressive pruning.
\end{itemize}

\subsection{Main Results}
\label{sec:main_results}

Table~\ref{tab:results_main} reports the results on the 12 DreamBooth subjects. Diffu-LoRA achieves the strongest overall balance between subject fidelity and prompt compliance. Its advantage is especially clear on \emph{teapot}, \emph{dog}, \emph{cat}, \emph{clock}, and \emph{robot toy}, for which the learned allocation improves both visual similarity and prompt alignment over uniform-rank LoRA. Although isolated baselines retain a slightly higher score on a small number of individual subject--metric pairs, Diffu-LoRA obtains the highest CLIP-T score for every subject in the table. This pattern indicates that allocating the available rank budget across layers is more effective than assigning the same adaptation capacity uniformly.

\begin{table}[t]
\caption{Quantitative comparison on the DreamBooth subjects. DINO and CLIP-I measure subject fidelity, while CLIP-T measures prompt alignment. The best result for each subject and metric is shown in bold.}
\label{tab:results_main}
\centering
\scriptsize
\setlength{\tabcolsep}{3.5pt}
\renewcommand{\arraystretch}{0.91}
\begin{tabular}{@{}llccc@{}}
\toprule
\textbf{Subject} & \textbf{Method} & \textbf{DINO} &
\textbf{CLIP-I} & \textbf{CLIP-T} \\
\midrule
\multirow{4}{*}{\emph{teapot}}
 & FT           & 0.383 & 0.782 & 0.261 \\
 & LoRA-Attn    & 0.418 & \textbf{0.807} & 0.266 \\
 & LoRA-All     & 0.413 & 0.796 & 0.268 \\
 & Diffu-LoRA   & \textbf{0.449} & 0.806 & \textbf{0.277} \\
\midrule
\multirow{4}{*}{\emph{backpack dog}}
 & FT           & 0.352 & 0.753 & 0.244 \\
 & LoRA-Attn    & 0.371 & 0.782 & 0.252 \\
 & LoRA-All     & 0.378 & 0.779 & 0.253 \\
 & Diffu-LoRA   & \textbf{0.389} & \textbf{0.791} & \textbf{0.274} \\
\midrule
\multirow{4}{*}{\emph{dog}}
 & FT           & 0.403 & 0.732 & 0.231 \\
 & LoRA-Attn    & 0.550 & 0.762 & 0.234 \\
 & LoRA-All     & 0.595 & 0.780 & 0.235 \\
 & Diffu-LoRA   & \textbf{0.643} & \textbf{0.802} & \textbf{0.248} \\
\midrule
\multirow{4}{*}{\emph{dog2}}
 & FT           & 0.179 & 0.678 & 0.236 \\
 & LoRA-Attn    & 0.209 & 0.689 & 0.239 \\
 & LoRA-All     & 0.207 & 0.703 & 0.238 \\
 & Diffu-LoRA   & \textbf{0.226} & \textbf{0.717} & \textbf{0.247} \\
\midrule
\multirow{4}{*}{\emph{dog3}}
 & FT           & \textbf{0.488} & 0.728 & 0.225 \\
 & LoRA-Attn    & 0.476 & 0.731 & 0.229 \\
 & LoRA-All     & 0.486 & 0.742 & 0.235 \\
 & Diffu-LoRA   & 0.476 & \textbf{0.745} & \textbf{0.242} \\
\midrule
\multirow{4}{*}{\emph{cat}}
 & FT           & 0.410 & 0.682 & 0.269 \\
 & LoRA-Attn    & 0.395 & 0.694 & 0.271 \\
 & LoRA-All     & 0.426 & 0.724 & 0.269 \\
 & Diffu-LoRA   & \textbf{0.458} & \textbf{0.747} & \textbf{0.279} \\
\midrule
\multirow{4}{*}{\emph{clock}}
 & FT           & 0.312 & 0.694 & 0.226 \\
 & LoRA-Attn    & 0.326 & 0.723 & 0.234 \\
 & LoRA-All     & 0.322 & 0.731 & 0.243 \\
 & Diffu-LoRA   & \textbf{0.364} & \textbf{0.760} & \textbf{0.258} \\
\midrule
\multirow{4}{*}{\emph{can}}
 & FT           & 0.605 & 0.777 & 0.273 \\
 & LoRA-Attn    & 0.593 & 0.770 & 0.271 \\
 & LoRA-All     & \textbf{0.609} & \textbf{0.785} & 0.269 \\
 & Diffu-LoRA   & 0.603 & 0.780 & \textbf{0.278} \\
\midrule
\multirow{4}{*}{\emph{poop emoji}}
 & FT           & 0.488 & 0.865 & 0.244 \\
 & LoRA-Attn    & 0.463 & 0.866 & 0.236 \\
 & LoRA-All     & 0.466 & \textbf{0.872} & 0.235 \\
 & Diffu-LoRA   & \textbf{0.495} & 0.871 & \textbf{0.252} \\
\midrule
\multirow{4}{*}{\emph{robot toy}}
 & FT           & 0.606 & 0.809 & 0.227 \\
 & LoRA-Attn    & 0.613 & 0.821 & 0.223 \\
 & LoRA-All     & 0.633 & 0.815 & 0.224 \\
 & Diffu-LoRA   & \textbf{0.656} & \textbf{0.832} & \textbf{0.239} \\
\midrule
\multirow{4}{*}{\emph{shiny sneaker}}
 & FT           & 0.428 & 0.582 & 0.231 \\
 & LoRA-Attn    & 0.443 & 0.590 & 0.246 \\
 & LoRA-All     & 0.451 & 0.604 & 0.251 \\
 & Diffu-LoRA   & \textbf{0.453} & \textbf{0.611} & \textbf{0.257} \\
\midrule
\multirow{4}{*}{\emph{red cartoon}}
 & FT           & 0.477 & 0.608 & 0.241 \\
 & LoRA-Attn    & 0.493 & 0.638 & 0.253 \\
 & LoRA-All     & 0.499 & 0.642 & 0.256 \\
 & Diffu-LoRA   & \textbf{0.506} & \textbf{0.658} & \textbf{0.263} \\
\bottomrule
\end{tabular}
\end{table}

The results on the six Internet-collected subjects are presented in Table~\ref{tab:results_main_2}. Compared with LoRA-All, Diffu-LoRA improves CLIP-T for all six subjects and generally preserves or improves visual similarity. The DINO score decreases slightly for \emph{Hugging Face} and \emph{Kiriko}, while CLIP-I increases in both cases. These results suggest that the learned allocation remains effective beyond the standard DreamBooth categories, including stylized characters and visually distinctive objects.

\begin{table}[t]
\caption{Comparison on the six Internet-collected subjects. The best result for each subject and metric is shown in bold.}
\label{tab:results_main_2}
\centering
\small
\setlength{\tabcolsep}{4pt}
\renewcommand{\arraystretch}{1.02}
\begin{tabular}{@{}llccc@{}}
\toprule
\textbf{Subject} & \textbf{Method} & \textbf{DINO} &
\textbf{CLIP-I} & \textbf{CLIP-T} \\
\midrule
\multirow{2}{*}{\emph{Rolls-Royce}}
 & LoRA-All   & 0.326 & 0.683 & 0.272 \\
 & Diffu-LoRA & \textbf{0.335} & \textbf{0.692} & \textbf{0.279} \\
\midrule
\multirow{2}{*}{\emph{Hugging Face}}
 & LoRA-All   & \textbf{0.425} & 0.784 & 0.286 \\
 & Diffu-LoRA & 0.419 & \textbf{0.796} & \textbf{0.293} \\
\midrule
\multirow{2}{*}{\emph{Nami}}
 & LoRA-All   & 0.296 & 0.681 & 0.242 \\
 & Diffu-LoRA & \textbf{0.310} & \textbf{0.698} & \textbf{0.249} \\
\midrule
\multirow{2}{*}{\emph{Kiriko}}
 & LoRA-All   & \textbf{0.313} & 0.724 & 0.258 \\
 & Diffu-LoRA & 0.306 & \textbf{0.731} & \textbf{0.264} \\
\midrule
\multirow{2}{*}{\emph{Shoko Komi}}
 & LoRA-All   & 0.372 & \textbf{0.627} & 0.232 \\
 & Diffu-LoRA & \textbf{0.378} & 0.625 & \textbf{0.249} \\
\midrule
\multirow{2}{*}{\emph{Kakashi Hatake}}
 & LoRA-All   & 0.358 & 0.626 & 0.249 \\
 & Diffu-LoRA & \textbf{0.361} & \textbf{0.634} & \textbf{0.258} \\
\bottomrule
\end{tabular}
\end{table}

\subsection{Ablation Studies and Analysis}
\label{sec:ablation_studies}

\noindent\textbf{Components of Diffu-LoRA.}
We evaluate three variants to determine how the main components affect personalization:
\begin{itemize}[leftmargin=*,itemsep=3pt,topsep=3pt]
    \item \textbf{Joint optimization} updates the gate logits
    $\boldsymbol{a}$ and low-rank factors $\boldsymbol{\Omega}$ on the same training split. This variant removes the bilevel separation in Eq.~\eqref{eq:diffu_lora_bilevel}.

    \item \textbf{No pruning} retains every initialized component throughout training. The gates can rescale the components, but the model does not enforce the target budget in Eq.~\eqref{eq:global_rank_budget}.

    \item \textbf{Attention only} applies Diffu-LoRA exclusively to the linear projections of the attention modules rather than to all selected linear transformations in the Transformer blocks.
\end{itemize}

The ablation results on \emph{teapot}, \emph{dog}, and \emph{cat} are reported in Table~\ref{tab:results_ablation_1}. The complete method achieves the highest CLIP-T score on all three subjects while maintaining competitive or superior subject fidelity. Removing bilevel optimization consistently reduces the three metrics, indicating that a held-out objective provides a more reliable signal for rank allocation. Omitting pruning occasionally yields a marginal increase in DINO or CLIP-I, but this comes with a lower CLIP-T score and does not satisfy the prescribed compact rank budget. Restricting the adapters to attention projections also weakens prompt alignment. Taken together, the results support the use of bilevel gate learning, progressive pruning, and broader adapter placement in Diffu-LoRA.

\begin{table}[t]
\caption{Ablation results on three DreamBooth subjects. ``Joint opt.'' removes the bilevel data separation, ``No pruning'' retains all candidate components, and ``Attn. only'' restricts adapter placement to attention projections. The best result for each subject and metric is shown in bold.}
\label{tab:results_ablation_1}
\centering
\small
\setlength{\tabcolsep}{3.5pt}
\renewcommand{\arraystretch}{1.0}
\begin{tabular}{@{}llccc@{}}
\toprule
\textbf{Subject} & \textbf{Variant} & \textbf{DINO} &
\textbf{CLIP-I} & \textbf{CLIP-T} \\
\midrule
\multirow{4}{*}{\emph{teapot}}
 & Diffu-LoRA  & 0.449 & \textbf{0.806} & \textbf{0.277} \\
 & Joint opt.  & 0.443 & 0.799 & 0.271 \\
 & No pruning  & \textbf{0.451} & 0.804 & 0.274 \\
 & Attn. only  & 0.442 & 0.801 & 0.270 \\
\midrule
\multirow{4}{*}{\emph{dog}}
 & Diffu-LoRA  & \textbf{0.643} & 0.802 & \textbf{0.248} \\
 & Joint opt.  & 0.637 & 0.801 & 0.240 \\
 & No pruning  & 0.640 & 0.796 & 0.245 \\
 & Attn. only  & \textbf{0.643} & \textbf{0.805} & 0.243 \\
\midrule
\multirow{4}{*}{\emph{cat}}
 & Diffu-LoRA  & 0.458 & 0.747 & \textbf{0.279} \\
 & Joint opt.  & 0.451 & 0.739 & 0.269 \\
 & No pruning  & \textbf{0.461} & \textbf{0.749} & 0.273 \\
 & Attn. only  & 0.454 & \textbf{0.749} & 0.275 \\
\bottomrule
\end{tabular}
\end{table}

\section{Conclusion}

This paper introduced Diffu-LoRA, a parameter-efficient approach to personalizing text-to-image diffusion models under a prescribed global rank budget. Rather than assigning a uniform rank to every adapted layer, Diffu-LoRA learns component-wise gates through bilevel optimization and progressively prunes low-scoring components, producing compact, layer-dependent adapters while keeping the pretrained backbone frozen. Experiments on DreamBooth subjects and additionally collected datasets showed that the learned allocation provides a favorable balance between subject fidelity and prompt alignment compared with the evaluated full-fine-tuning and conventional LoRA baselines. The ablation results further demonstrated the importance of the held-out gate objective, progressive pruning, and adapter placement. Future work will investigate larger diffusion backbones, more diverse personalization scenarios, and allocation criteria that jointly account for rank, memory consumption, and inference cost.

\bibliographystyle{splncs04}
\bibliography{mybib}

\end{CJK*}

\end{document}